\documentclass[runningheads]{llncs}
\usepackage{makeidx}
\usepackage{graphicx}
\usepackage{amsmath,amssymb} 
\usepackage{color}
\usepackage{subfig}
\usepackage{array}

\def\etal{et\,al.}
\newcolumntype{P}[1]{>{\centering\arraybackslash}p{#1}}
\newcommand{\Realnumbers}{\mathrm{I \! R}}

\begin{document}
	\pagestyle{headings}
	\mainmatter

	\title{SDNet: Semantically Guided \\ Depth Estimation Network}

	\titlerunning{SDNet: Semantically Guided Depth Estimation Network}
	\authorrunning{M. Ochs, A. Kretz, and R. Mester}
\author{
Matthias Ochs\inst{1} \and
Adrian Kretz\inst{1} \and
Rudolf Mester\inst{1,2}
}

\institute{
$^1$ VSI Lab, Goethe University, Frankfurt am Main, Germany \\
$^2$ Norwegian Open AI Lab, CS Dept. (IDI), NTNU Trondheim, Norway
}

\maketitle

\begin{abstract}

Autonomous vehicles and robots require a full scene understanding of the environment to interact with it. Such a perception typically incorporates pixel-wise knowledge of the depths and semantic labels for each image from a video sensor. Recent learning-based methods estimate both types of information independently using two separate CNNs. In this paper, we propose a model that is able to predict both outputs simultaneously, which leads to improved results and even reduced computational costs compared to independent estimation of depth and semantics. We also empirically prove that the CNN is capable of learning more meaningful and semantically richer features. Furthermore, our SDNet estimates the depth based on ordinal classification. On the basis of these two enhancements, our proposed method achieves state-of-the-art results in semantic segmentation and depth estimation from single monocular input images on two challenging datasets.

\end{abstract}
\section{Introduction}

Many computer vision applications rely on or at least can increase their performance, if pixel-wise depth information is available in addition to the input image. Based on this depth information, geometric relationships in the environment can be explained better and can support, for instance, the scene understanding in robotics, autonomous driving, image editing or 3D modeling/reconstruction.

\begin{figure}[ht]
	\centering
	\includegraphics[width=1.0\linewidth]{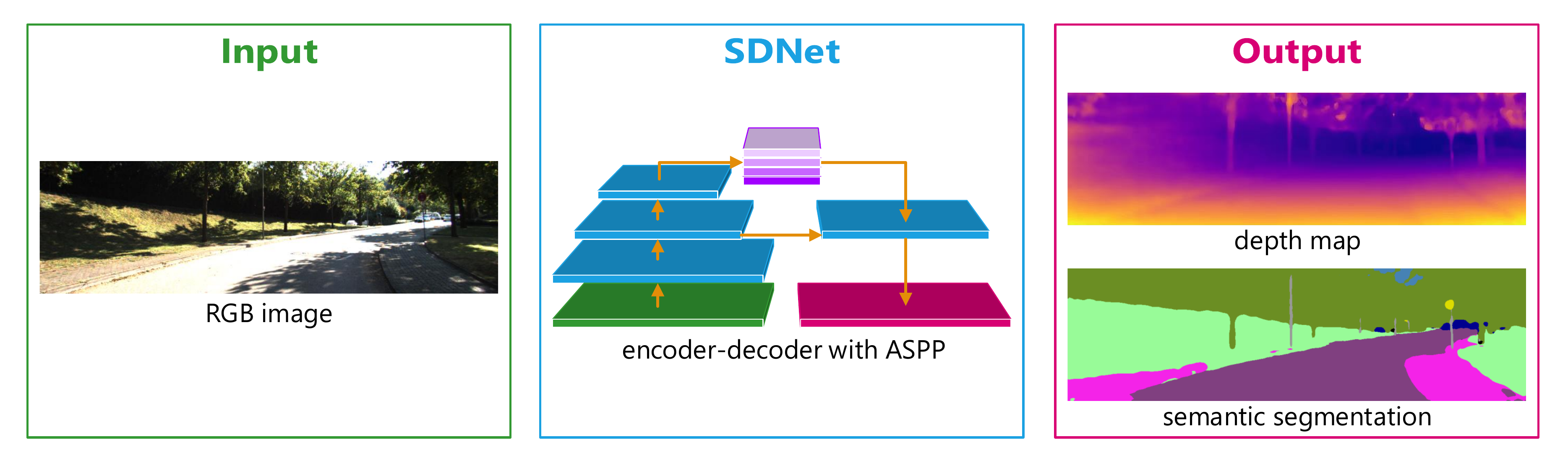}
	\caption{Our proposed SDNet estimates pixel-wise depth and semantic labels from a single monocular input image based on a modified DeepLabv3+ architecture.}
	\label{intro:overview}
\end{figure}

Unfortunately, estimating the depth from a single monocular image is an ill-posed problem, which cannot be solved without further knowledge. In this paper, we propose a deep learning-based approach, where this additional information is learned in a supervised manner from the geometric cues of many training samples by a CNN. The network architecture of this CNN is inspired by the DeepLabv3+ model from \cite{Chen2018} and extended by two subnets to simultaneously predict the depth and semantic labels. Due to this combined prediction, our net learns semantically richer and more stable features in the Atrous Spatial Pyramid Pooling (ASPP) module, which results in an improved estimation of both. Following the insights of \cite{Fu2018} and having in mind that depths follow a natural order, we regard the depth estimation as an ordinal classification problem and not as a regression-based one. This ensures that the neural network converges faster and that the resulting depth maps are more precise and can retrieve more details of fine structures and small objects. In the evaluation section, we show that SDNet yields state-of-the-art results on two datasets while still being able to generalize to other datasets as well.
\section{Related Work}

In this section, we review classical as well as deep learning based methods for solving the depth estimation problem. In most classical approaches \cite{Saxena2005,Saxena2009,Liu2014}, hand-crafted features are extracted from the monocular images, which are then used to predict the depth map by optimizing a probabilistic model like, a Markov random field (MRF) or conditional random field (CRF). Ladicky \etal \cite{Ladicky2014} estimate the depth based on different canonical views and show that semantic knowledge helps to improve the prediction. This is verified for stereo-matching methods by \cite{Schneider2016a,Yang2018a}, too. In this paper, we empirically prove that this concept also holds for depth estimation from a single monocular image with a CNN.

Besides these classical techniques, many recent approaches make use of Convolutional Neural Networks (CNNs). These techniques can be divided into two categories: supervised and unsupervised learning methods. Supervised learning techniques include models which are trained on stereo images, but which can infer depth maps on monocular images. \cite{Knoebelreiter2017} propose an approach which follows this paradigm. They use the correlations of CNN feature maps of stereo images and derive the unary matching costs. The depth maps are then computed by using a CRF. But there are also CNNs \cite{Zbontar2016,Mayer2016,Pang2017,Gidaris2017}, which try to find stereo correspondences to estimate the disparity/depth.

The previously mentioned approaches are trained in a supervised manner. \cite{Kuznietsov2017} introduce a semi-supervised approach, where the loss function includes an additional unsupervised term. This term captures the stereo alignment error of two images given an approximate depth map. The approaches of \cite{Garg2016,Xie2016} and \cite{Godard2017} use modifications of this alignment error to train networks in a completely unsupervised way. Since ground truth depth maps are not necessary with these approaches, it is possible to make use of much larger training databases. These unsupervised techniques tend to produce much smoother and thus more inaccurate depth map than supervised techniques, though.

Fu \etal \cite{Fu2018} introduced DORN, which is a CNN trained in a supervised way and achieves current state-of-the-art results. Instead of using regression to estimate the depth of each pixel, the depths are discretized and ordinal classification is used. We also use a similar idea for our depth estimation. 

\section{Approach}

In this section, we introduce our proposed model, which classifies the depth based on ordinal depth classes while simultaneously inferring the semantic labels for each pixel, too. This has the advantage that more meaningful features are learned in the final layers of the encoder, from which both the semantic segmentation and the depth estimation can benefit. This allows the CNN to describe, detect and classify objects more accurately.

\subsection{Architecture} 
\label{approach:architecture}

\begin{figure}[ht]
	\centering
	\includegraphics[width=1.0\linewidth]{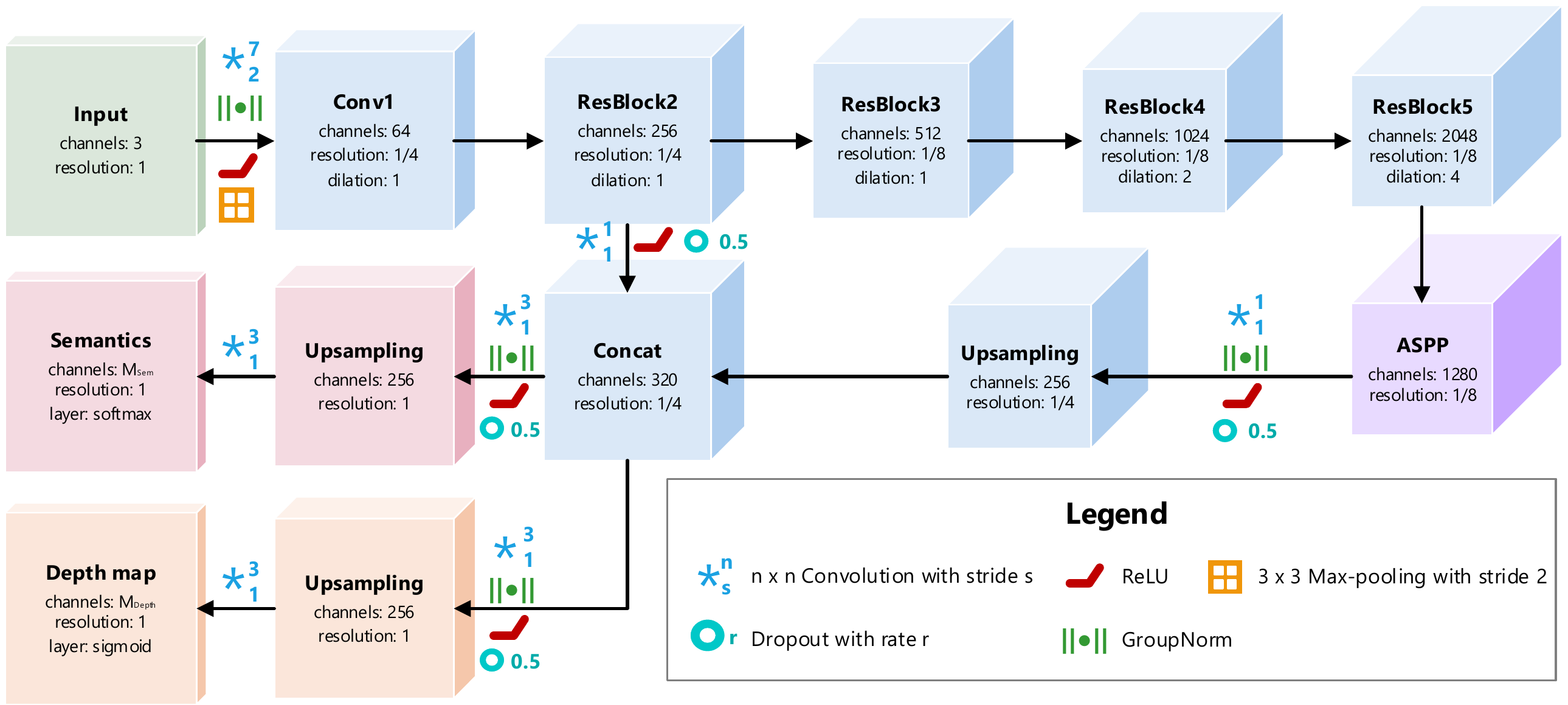}
	\caption{Encoder-decoder architecture of SDNet with ASPP module. The decoder consists of two subnets, which predict the pixel-wise depths and semantic labels.}
	\label{fig:approach:architecture}
\end{figure}

The network architecture of our model is depicted in figure~\ref{fig:approach:architecture}. It is similar to the DeepLabv3+ model from \cite{Chen2018}, which represents the current state of the art for semantic segmentation. 

Theoretically, every encoder, e.g.\ ResNets or InceptionNets, can be used to extract feature maps which serve as input for the ASPP module. Due to limited computational capacity and real-time requirements, we use a ResNet-50 in this work. The ResNet is configured as proposed by \cite{Chen2018} (dilation rate, multi grid, output stride of $ 8 $, atrous convolution, etc.). With the combination of this encoder and the ASPP module, SDNet is capable of extracting features on multiple scales, which improves the results significantly. The ASPP module is also adopted from DeepLabv3+, except that we use group normalization \cite{WuHe2018} instead of batch normalization, which applies to all normalization layers in the CNN. Furthermore, we add dropout layers, as shown in figure~\ref{fig:approach:architecture}. Compared to DORN \cite{Fu2018}, our ASPP module is \textit{fully-convolutional} and thus our CNN is not dependent on a predefined input resolution.

Unlike DeepLabv3+, our decoder consists of two subnets. First, the decoder interpolates the semantically meaningful feature maps from the ASPP module which are then concatenated with feature maps from the second ResBlock. These feature maps provide additional structural information and are able to recognize fine structures in the image. The resulting feature maps serve as input for our two subnets. In the first subnet, the semantic labels are estimated. The second subnet determines the depth. Both output layers of the two subnets compute class probabilities, which is done by a softmax function for the semantics and by a sigmoid layer for the depths.

\subsection{Discretization}

For ordinal regression, the continuous depth must be discretized so that a particular class $ c $ can be assigned to each depth $ d $. Normally, the depths $ d \in [d_{\text{min}}, \: d_{\text{max}}] $  are divided linearly into classes between the minimum and maximum depth. However, this partitioning has the disadvantage that an erroneous estimation of a depth class at shallow depths causes a larger relative estimation error compared to a faulty estimated large depth.

\begin{figure}[ht]
	\centering
	\includegraphics[width=.49\linewidth]{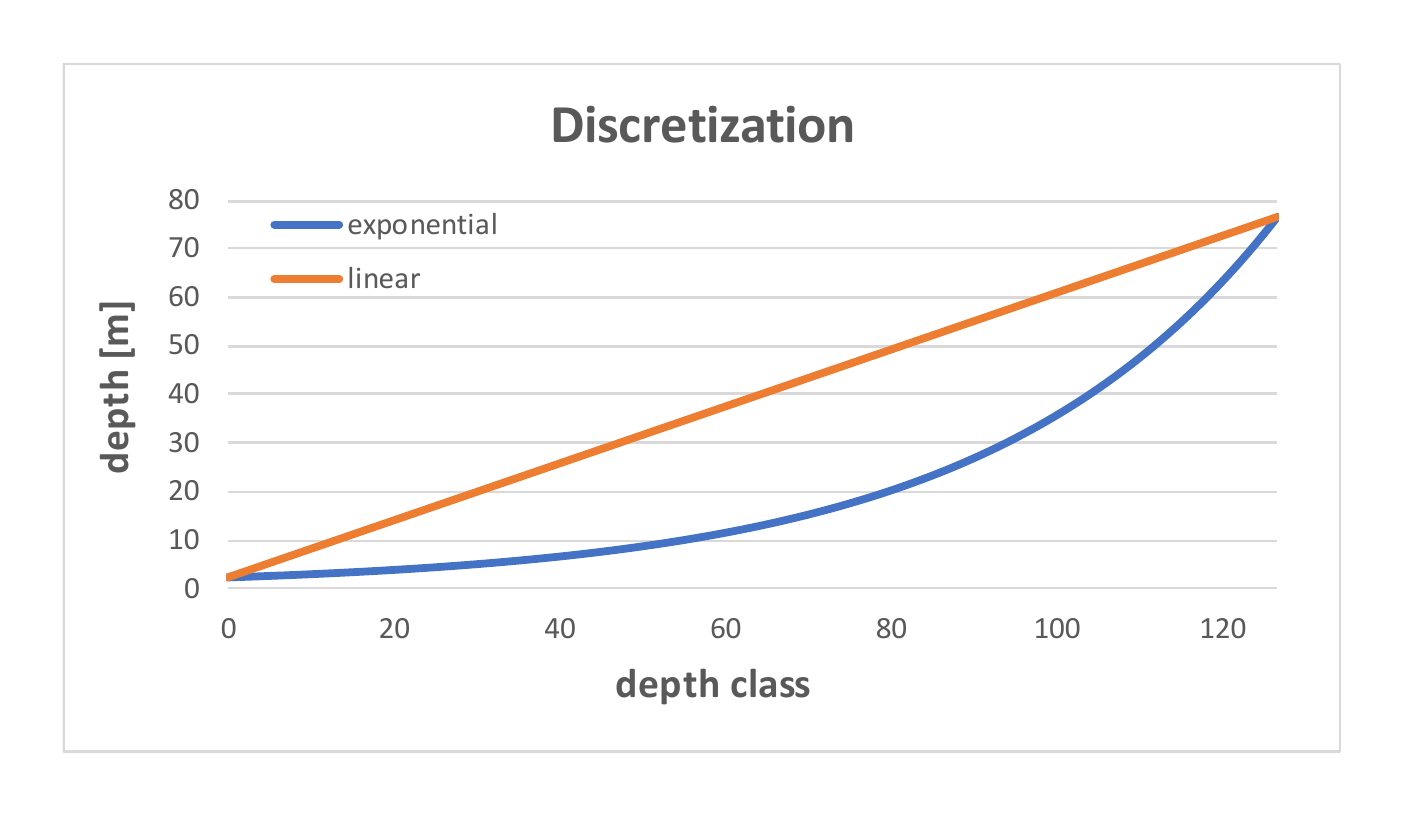}
	\includegraphics[width=.49\linewidth]{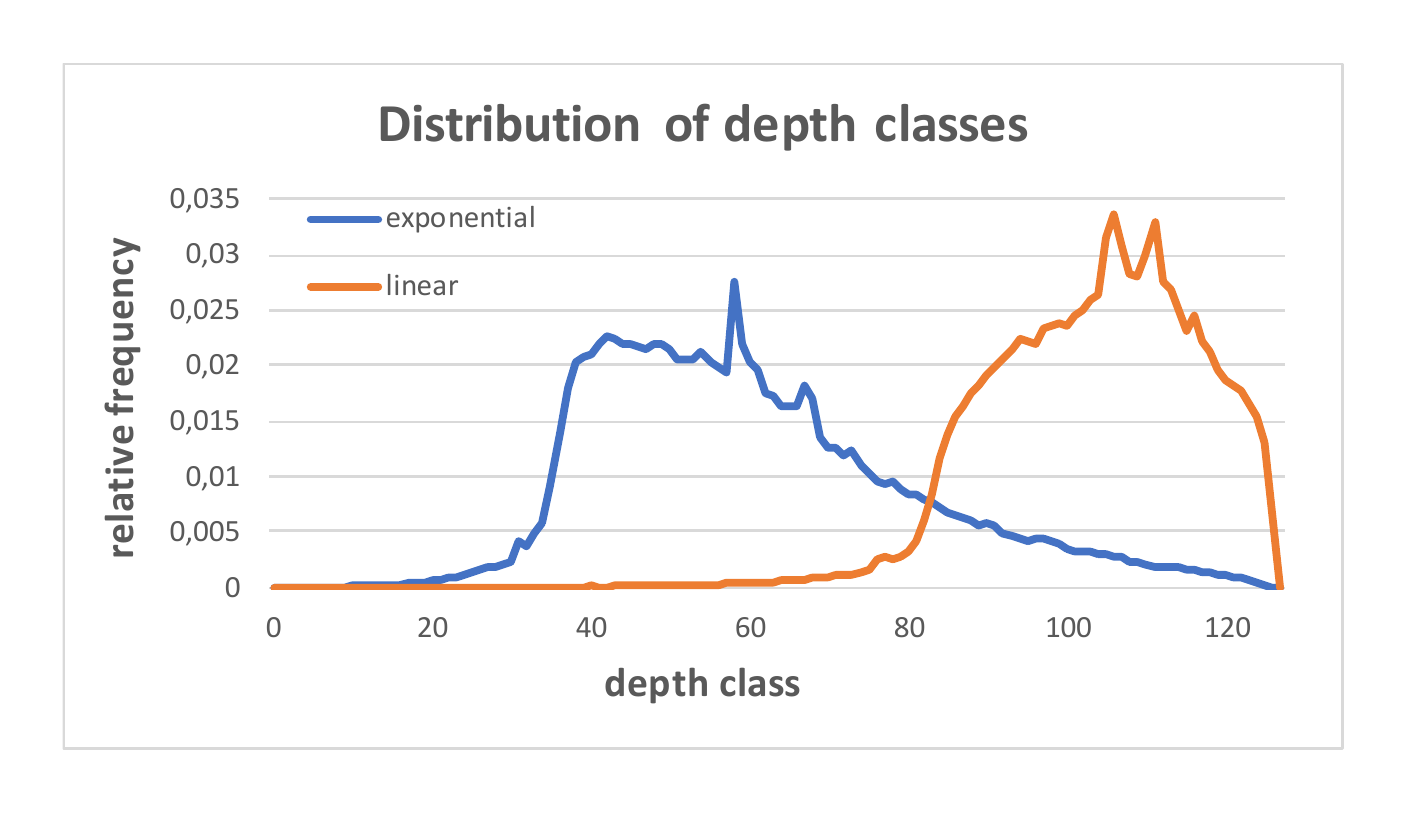}
	\caption{The left plot shows the exponential and linear discretization of the depths with $ M_{\text{depth}} = 128 $ in the interval $ [2 \text{ m}, \: 80 \text{ m}] $. In the right histogram, the relative frequency of these exponential and linear discretized depth classes are shown over all existing depths of the KITTI depth prediction training dataset.}
	\label{fig:approach:discretization}
\end{figure}

This relative error can be reduced by either defining the classes by their inverse depths $ \rho = \frac{1}{d} $, or by choosing a more complex mapping function, e.g., \cite{Fu2018} chose a logarithmic mapping function for discretization. In this work, we split the inverse depth into discrete classes $ c_i $ using an exponential function
\begin{equation}
	c_i = \rho_{\text{min}} \cdot \exp\left(\log\left(\frac{\rho_{\text{\tiny{max}}}}{ \rho_{\text{\tiny{min}}} }\right) \cdot \frac{i - 1}{M_{\text{\tiny{d}}} - 1}\right), \quad i \in \{1, \ldots M_{\text{d}}\}
\end{equation}
where $ M_{\text{d}} $ corresponds to the number of depth classes. In the left plot of figure~\ref{fig:approach:discretization}, both the linear and the exponential mapping functions are shown. On the basis of this graph, one can see that due to the exponential character, more discretization levels or classes are available for small and medium depths than for larger ones. Thus, the relative prediction errors of an estimator should be correspondingly smaller. In the adjacent histogram, the relative frequencies of these two depth classes are shown using the depths found in the KITTI depth prediction training data. A uniform distribution of these classes would be ideal for training a CNN. By a suitable choice of the mapping function, such as the proposed exponential, the depth distribution can be approximated to be a uniform one.

In contrast to the semantic classes, the depth classes follow an ordinal relation. This means that a larger depth class $ c_i $ is always included in the smaller ones: $ \{ c_1 \preccurlyeq c_2 \preccurlyeq \ldots c_i \ldots \preccurlyeq c_{M\text{\tiny{d}}} \} $. By this concept, an inverse depth $ \rho $ which actually belongs to the class $ c_i $, also belongs to the classes $ c_1, \ldots, c_{i-1} $. This ordered classification was introduced for neural networks by \cite{Cheng2008} and \cite{Niu2016}. \cite{Fu2018} and we use this idea, and show that an ordinal classification is advantageous over a standard classification or regression based estimation of depth values.

\subsection{Output \& Loss}

As described in Section~\ref{approach:architecture}, SDNet returns the result of the semantic segmentation and depth estimation in two separate subnets. The output and the loss of the semantic subnet follow the standard multi-class classification for each pixel $ k \in {1, \ldots, K} $, where $ K $ is the number of pixels in the image. Hence, the semantic subnet outputs logits $ \vec{y}_k^s \in \Realnumbers^{M_s} $, where $M_s $ is equal to the number of semantic classes. The logits $ \vec{y}_k $ are then converted into probabilities $ \vec{p}_k^s $ by using the softmax function. As training loss, we use the multi-class cross entropy with $ \vec{t}_k^s \in \{0,1\}^{M_s} $ as the one-hot-encoded vector representing the ground truth:
\begin{equation}
	L_{k, \text{sem}}(\vec{\theta}; \vec{p}_{k}^s, \vec{t}_{k}^s) = - \frac{1}{M_s} \cdot \sum_{i} t_{k,i}^s \log p_{k,i}^s,
\end{equation}
where $ \vec{\theta} $ are the trainable parameters in the CNN.

This vanilla multi-class classification cannot be used for the ordered depth classes, because a single pixel can belong to multiple labels (multi-label classification), whereas each depth class is not dependent on the others. Thus, it is regarded as a binary classification problem for each class $ c_i $ and pixel $ k $. The depth logits $\vec{y}_k^d \in \Realnumbers^{M _{\text{d}}} $ are transformed by the sigmoid function to probabilities $ \vec{p}_ k^d $, where $ p_{i, k}^d \in [0, 1] $ indicates whether the pixel $ k $ belongs to class $ c_i $ or not. Then, SDNet calculates the following multi-hot-encoded vector for each pixel:
\begin{equation}
	\begin{aligned}
	\vec{z}_k^d &= \left[\sigma(p^d_{k, 1}), \ldots, \sigma(p^d_{k, M_d}) \right]^T, \quad \text{with} \\
	\sigma(x) &= \begin{cases}
	1, & \text{if } x \geq 0.5 \\
	0, & \text{otherwise.} 
	\end{cases}
	\end{aligned}
\end{equation}

To preserve the ordered relation of the classes, the components of the classified vector $ \vec{y}_i $ are accumulated until the first entry of the vector $ \vec{z}_k^d $ is $ 0 $. This sum then corresponds to the class of the pixel:
\begin{equation}
	\begin{aligned}
	c_k &= \sum_{i} \eta(\vec{z}^d_k, i), \quad \text{with} \\
	\eta(\vec{z}^d_k, i) &= \begin{cases}
	1, & \text{if~} i = 1 \text{~and~} \vec{z}^d_{k,i} = 1 \\
	1, & \text{if~} \eta(\vec{z}^d_k, i-1) = 1 \text{~and~} \vec{z}^d_{k,i} = 1 \\
	0, & \text{otherwise.} 
	\end{cases}
	\end{aligned}
\end{equation}

To train the depth part of SDNet, we choose the following binary cross entropy loss and the ground truth vector $\vec{t}_k^d $ is formulated as a multi-hot-encoded vector for class $ c_k $:
\begin{equation}
	L_{k, \text{depth}}(\vec{\theta}; \vec{p}^d_k, \vec{t}^d_k) = - \frac{1}{M_{\text{d}}} \cdot \sum_{i} t_{k,i}^d \log p_{k,i}^d + (1-t_{k,i}^d) \log (1 - p_{k,i}^d).
	\label{eq:approach:bce_loss}
\end{equation}

Thus, the total loss is the mean of two individual losses over all pixels in the batch, with $ \lambda $ acting as a coupling constant:
\begin{equation}
	L(\vec{\theta}; \vec{p}_k, \vec{t}_k) = \frac{1}{K} \sum_{k} L_{k, \text{depth}}(\vec{\theta}; \vec{p}_k^d, \vec{t}_k^d) + \lambda \cdot L_{k, \text{sem}}(\vec{\theta}; \vec{p}_k^s, \vec{t}_k^s).
\end{equation}

To weight both loss terms similarly, $ \lambda =  10 $ was empirically determined. We have also analyzed additional loss terms, such as regularization of the predicted depth map by its gradient as proposed by \cite{Kuznietsov2017} and \cite{Godard2017}, or the absolute sum of the depth values from \cite{Yang2018}. Another possible improvement would be the use of class balancing factors or the focal loss from \cite{Lin2018}. However, all of these improvements have not produced any additional positive effects, so we disregarded them.

\section{Evaluation \& Experiments}

This section describes the training of SDNet in more detail. Both, the predicted depth and semantic segmentation are evaluated and achieve state-of-the-art results in the KITTI depth prediction (KD) benchmark \cite{Uhrig2017} and the KITTI pixel-level semantic segmentation (KS) benchmark \cite{Alhaija2018}. In addition, we demonstrate the generalization of SDNet to other data on Cityscapes (CS) \cite{Cordts2016} and compare our semantic segmentation results with a modified DeepLabv3+ model. We use the same evaluation metrics as in \cite{Eigen2014} and the official benchmarks.

Both the evaluation and ablation studies show that a joint estimation of depths and semantic labels is preferable to an independent estimation, because it yields superior results. This observation indicates that a CNN is learning semantically more meaningful features. Using these features, the CNN can detect and recognize objects better, which in turn improves both predictions. Additionally, the experiments also show that depths can be estimated more precisely using classes rather than a regression-based approach.

\subsection{Training Protocol}

To train the SDNet, we used the three datasets mentioned above in the following order. First, we train on the $3000$ samples of fine-annotated training data from CS, which contain both semantic information and depth maps from SGBM \cite {Hirschmueller2008}. These depth maps are relatively dense compared to LIDAR measurements from KITTI, but they are also less accurate and exhibit the typical stereo artifacts. Thus, SDNet was also trained with the $ 85898 $ samples of training data from KD, in which LIDAR measurements were accumulated over several frames to get denser depth maps. There is no semantic information available for this data, so only the depth part of SDNet could be trained with KD. However, KITTI published a training dataset (KS) with $ 200 $ images containing semantic annotations. LIDAR data also exists for this dataset, so that all necessary data is available for a complete training of SDNet.

Due to the different resolutions and the limited GPU memory, we randomly cropped the images to $ 768 \times 352$ pixel at each epoch. Furthermore, we used random flipping and the standard color jittering as data augmentation techniques. We initialized the encoder with pre-trained weights from ImageNet and froze the parameters of the first and the second ResNet block during training. We used Adam as optimizer and a batch size of $ 4 $. The weight-decay factor and dropout rate was set to $ 0.0005 $ and $ 0.5 $ respectively. The best results were achieved when using $ 128 $ depth classes to split the depth interval of $ [2 \text{m}, 80 \text{m}]$. 

In the first training phase, SDNet was trained on CS and KD with $50.000$ iterations each and an initial learning rate of $ 0.0001$. For training on KD, $ \lambda $ was set to $ 0 $ because there is no semantic data and thus only the depth subnet should be trained. In all other cases $ \lambda = 10 $ was chosen. The second training phase consists of fine-tuning on KS, where both subnets are optimized using learning rate of $ 0.00001$ and $ 3000 $ iterations.

During deployment, there is no need to divide the input images, because SDNet is fully-convolutional. Thus, the output can be estimated on arbitrary sized input images. Similarly, as suggested by \cite {Godard2017}, SDNet also predicts the output of the flipped input image. The final prediction is then computed by merging these two outputs, which makes the result more robust to occlusion and outliers.

\subsection{KITTI Dataset}

The depth estimation and semantic segmentation were evaluated by several evaluations within KD and KS, and compared to state-of-the-art methods. Before KITTI published its benchmark for the monocular depth estimation in $ 2018 $, it was common to evaluate the results using an unofficial dataset suggested by \cite{Eigen2014}. 
\begin{table}[ht]
	\begin{center}
		\begin{tabular}{l | P{1.1cm} P{1.1cm} P{1.2cm} P{1.6cm} | P{1.5cm} P{1.6cm} P{1.6cm}}
			Method & ARD $\downarrow$ & SRD $\downarrow$ & RMSE $\downarrow$ & $\text{RMSE}_{\text{Log}} \downarrow$ & $ \delta < 1.25 \uparrow $ & $ \delta < 1.25^2 \uparrow $ & $ \delta < 1.25^3 \uparrow $\\ 
			\hline
			\hline
			Fu \etal \cite{Fu2018}         & 0.072 & 0.307 & 2.727 & 0.120 & 0.932 & 0.985 & 0.995   \\
			\textbf{SDNet}        & 0.079 & 0.504 & 3.700 & 0.167 & 0.921 & 0.968 & 0.983  \\
			Yang \etal \cite{Yang2018}		  & 0.097 & 0.734 & 4.442 & 0.187 & 0.888 & 0.958 & 0.980  \\
			Kuznietsov \etal \cite{Kuznietsov2017} & 0.113 & 0.741 & 4.621 & 0.189 & 0.862 & 0.960 & 0.986  \\
			Godard \etal \cite{Godard2017}     & 0.114 & 0.898 & 4.935 & 0.206 & 0.861 & 0.949 & 0.976  \\
			Eigen \etal \cite{Eigen2014}      & 0.190 & 1.515 & 7.156 & 0.270 & 0.692 & 0.899 & 0.967  \\
			Liu \etal \cite{Liu2016}        & 0.217 & 1.841 & 6.986 & 0.289 & 0.647 & 0.882 & 0.961  \\
			Saxena \etal \cite{Saxena2009}     & 0.280 & 3.012 & 8.734 & 0.361 & 0.601 & 0.820 & 0.926  \\
			\hline
			Fu \etal \cite{Fu2018}         & 0.071 & 0.268 & 2.271 & 0.116 & 0.936 & 0.985 & 0.995   \\
			\textbf{SDNet}        & 0.075 & 0.430 & 3.199 & 0.163 & 0.926 & 0.970 & 0.984  \\
			Yang \etal \cite{Yang2018}       & 0.092 & 0.547 & 3.390 & 0.177 & 0.898 & 0.962 & 0.982  \\
			Kuznietsov \etal \cite{Kuznietsov2017} & 0.108 & 0.595 & 3.518 & 0.179 & 0.875 & 0.964 & 0.988  \\
			Godard \etal \cite{Godard2017}     & 0.108 & 0.657 & 3.729 & 0.194 & 0.873 & 0.954 & 0.979  \\
			Garg \etal \cite{Garg2016}       & 0.169 & 1.080 & 5.104 & 0.273 & 0.740 & 0.904 & 0.962  \\
			\hline
		\end{tabular}
	\end{center}
	\caption{Evaluation and comparison with other approaches on the test split by Eigen \etal \cite{Eigen2014}. In the upper part of this table, the depth is evaluated in range of $ [0 \text{m}; 80 \text{m}] $ whereas the maximum depth was reduced to $ 50 $ m in the lower part.}
	\label{table:experiments:kitti_eigen_split}
\end{table}
\begin{figure}[ht]
	\centering
	\captionsetup[subfigure]{position=top, labelformat=empty, farskip=2pt}
	\subfloat[RGB]{\includegraphics[width=.24\linewidth]{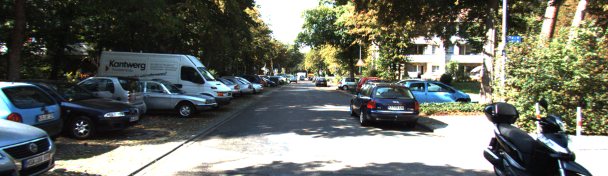}}
	\hspace{1pt}
	\subfloat[GT]{\includegraphics[width=.24\linewidth]{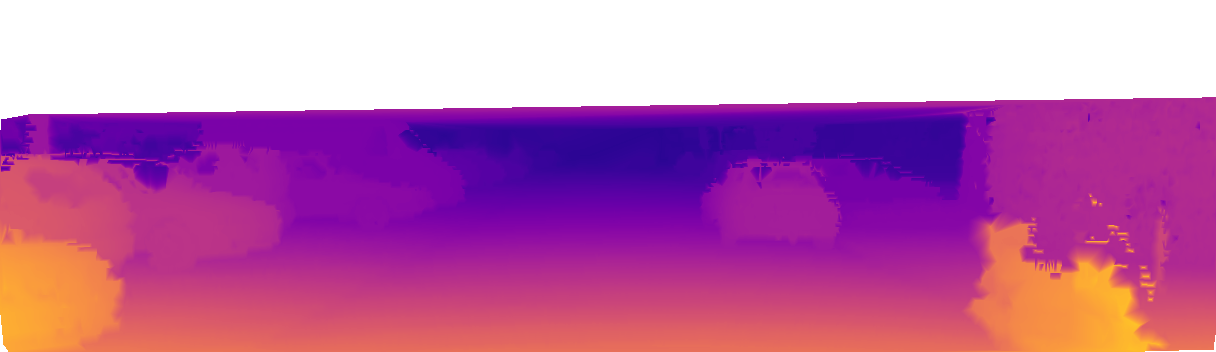}}
	\hspace{1pt} 
	\subfloat[Garg \etal]{\includegraphics[width=.24\linewidth]{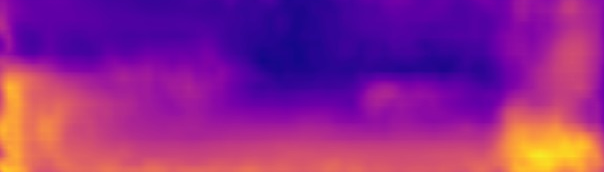}}
	\hspace{1pt} 
	\subfloat[Liu \etal]{\includegraphics[width=.24\linewidth]{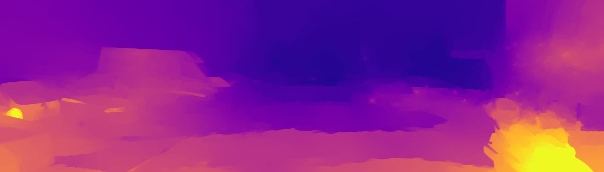}} \\
	\captionsetup[subfigure]{position=bottom}
	\subfloat[Godard \etal]{\includegraphics[width=.24\linewidth]{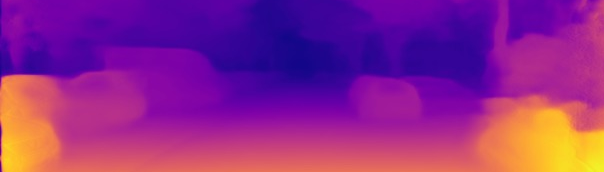}}
	\hspace{1pt}  
	\subfloat[Kuznietsov \etal]{\includegraphics[width=.24\linewidth]{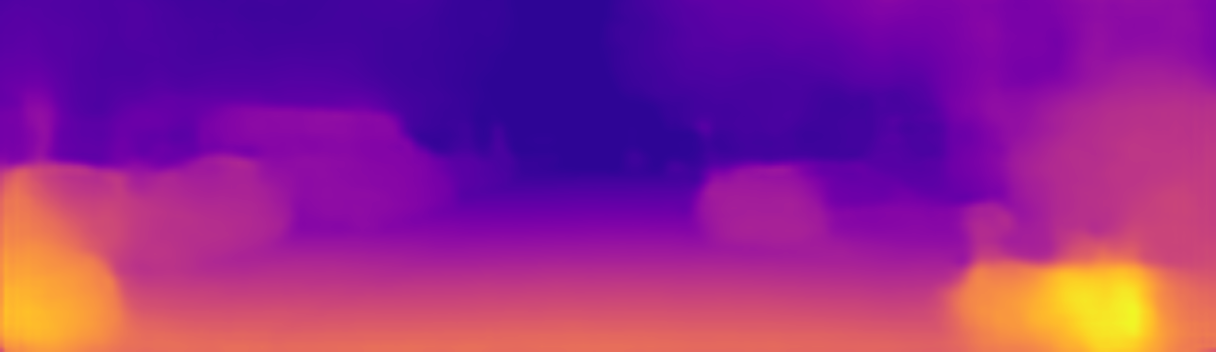}}
	\hspace{1pt}  
	\subfloat[Yang \etal]{\includegraphics[width=.24\linewidth]{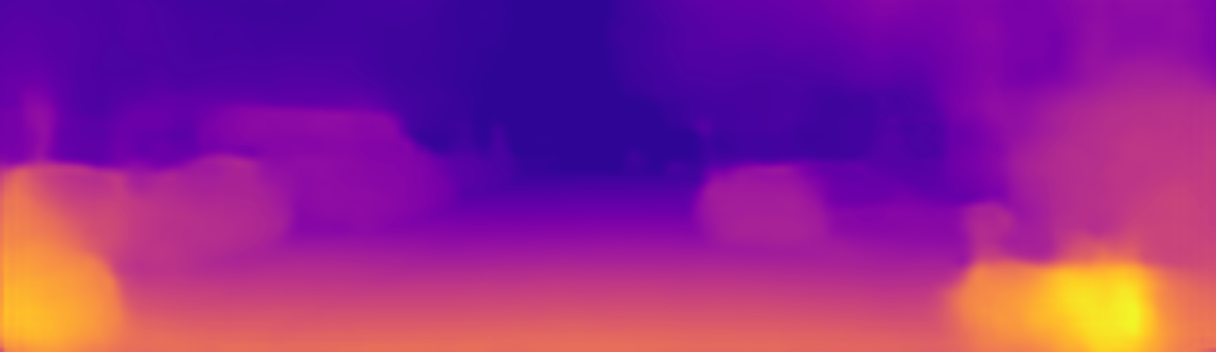}}
	\hspace{1pt}  
	\subfloat[SDNet]{\includegraphics[width=.24\linewidth]{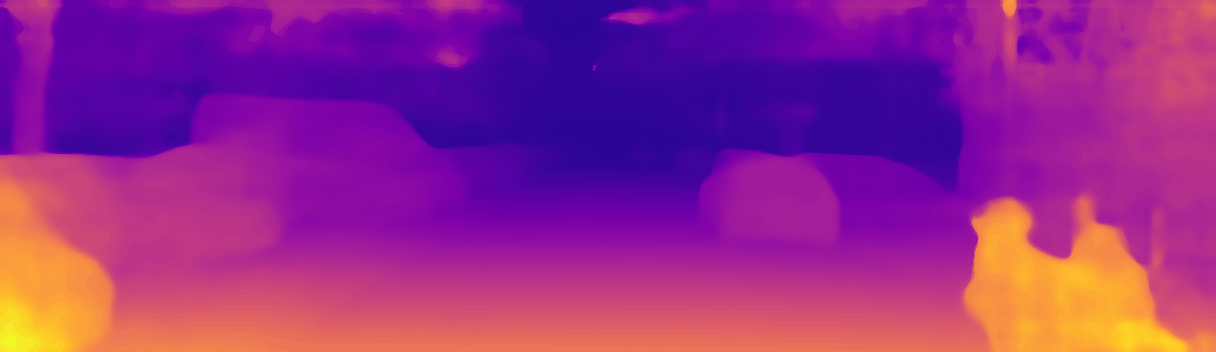}} \\
	\vspace{0.1cm}
	\subfloat{\includegraphics[width=.24\linewidth]{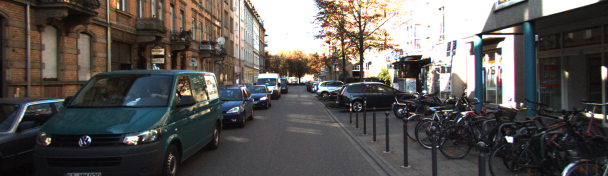}}
	\hspace{1pt}  
	\subfloat{\includegraphics[width=.24\linewidth]{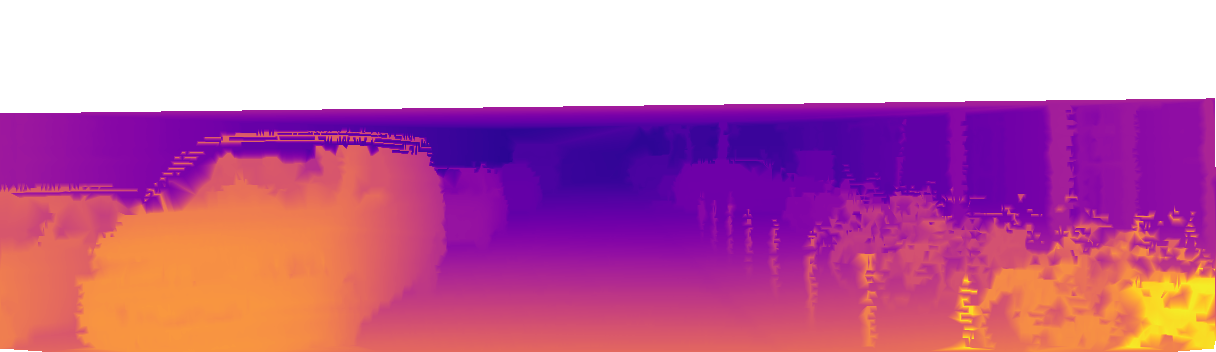}} 
	\hspace{1pt} 
	\subfloat{\includegraphics[width=.24\linewidth]{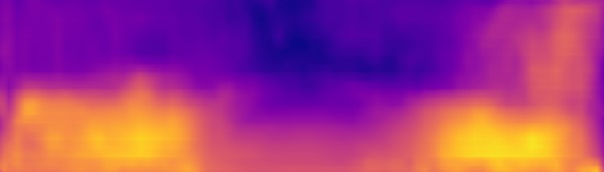}} 
	\hspace{1pt} 
	\subfloat{\includegraphics[width=.24\linewidth]{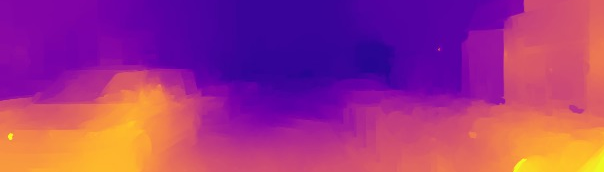}} \\
	\subfloat{\includegraphics[width=.24\linewidth]{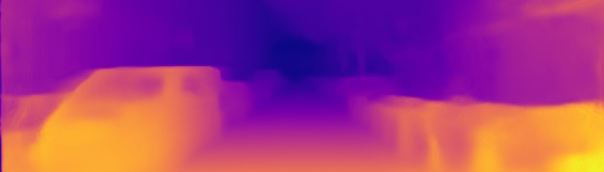}}\hspace{1pt} 
	\subfloat{\includegraphics[width=.24\linewidth]{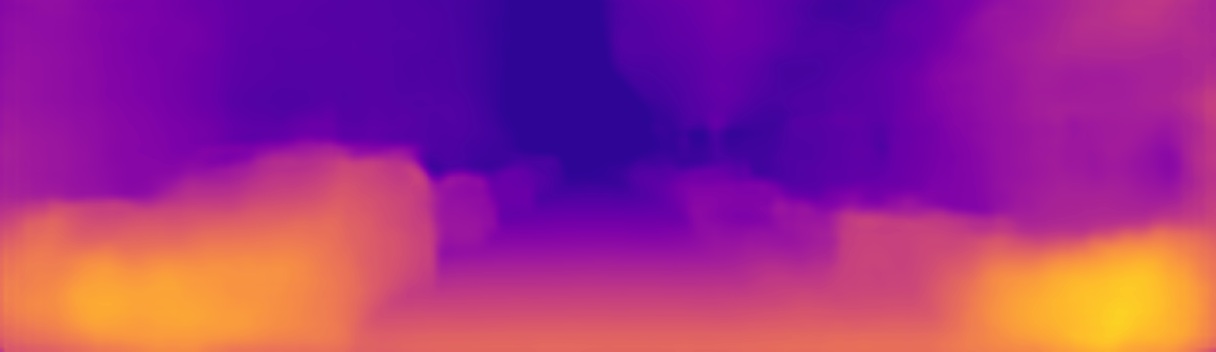}} 
	\hspace{1pt}
	\subfloat{\includegraphics[width=.24\linewidth]{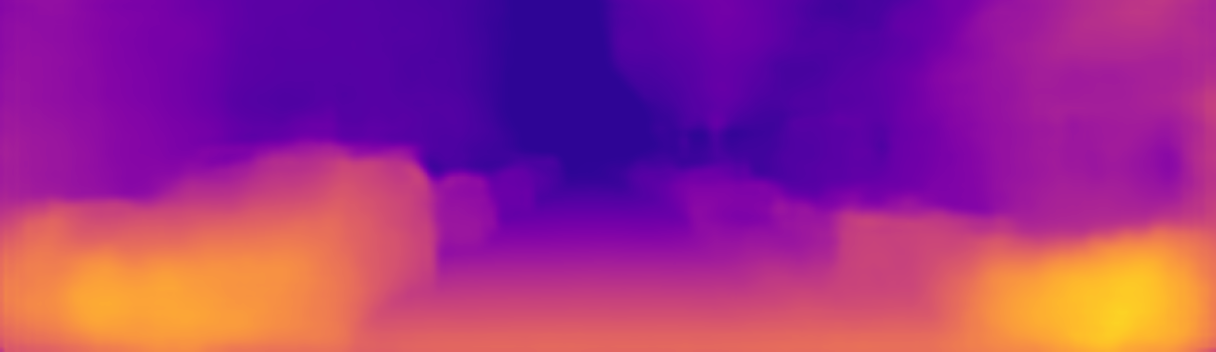}} 
	\hspace{1pt}
	\subfloat{\includegraphics[width=.24\linewidth]{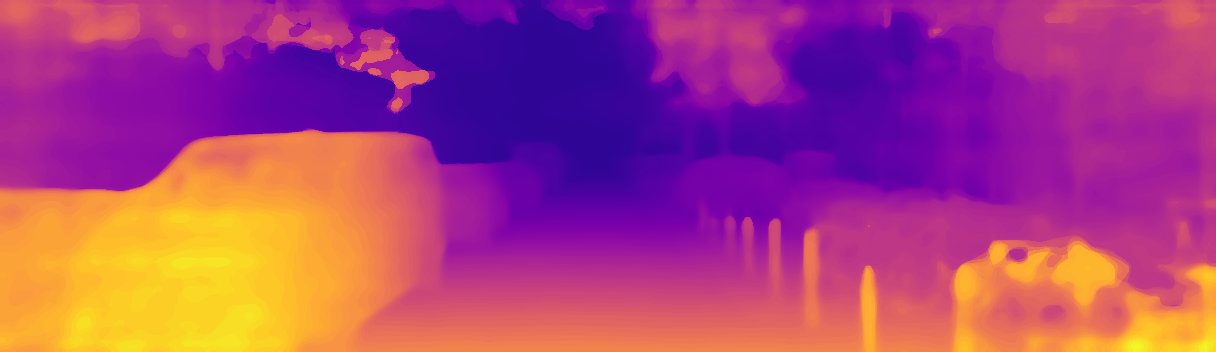}} \\
	\vspace{0.1cm}
	\subfloat{\includegraphics[width=.24\linewidth]{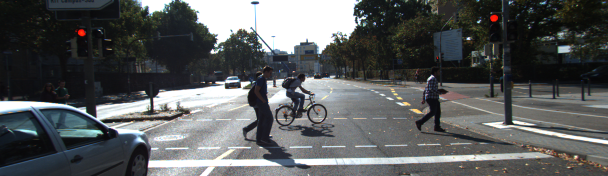}}
	\hspace{1pt} 
	\subfloat{\includegraphics[width=.24\linewidth]{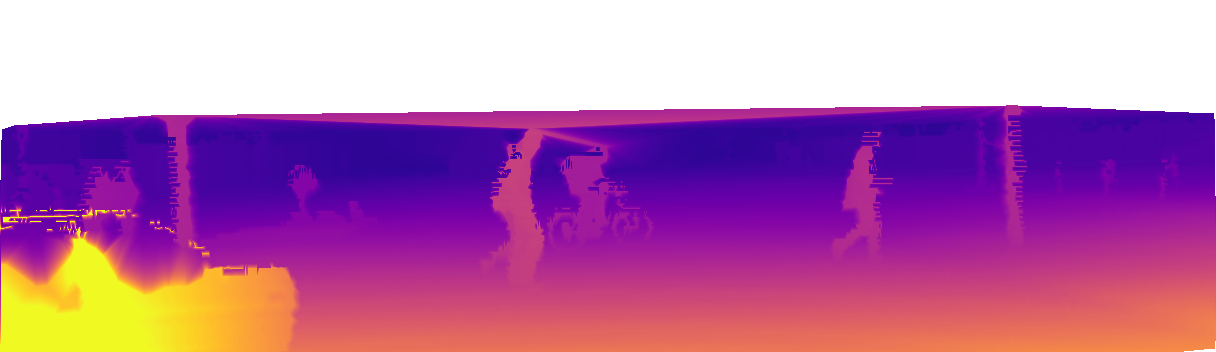}}
	\hspace{1pt} 
	\subfloat{\includegraphics[width=.24\linewidth]{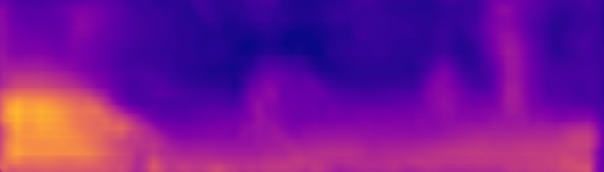}} 
	\hspace{1pt}
	\subfloat{\includegraphics[width=.24\linewidth]{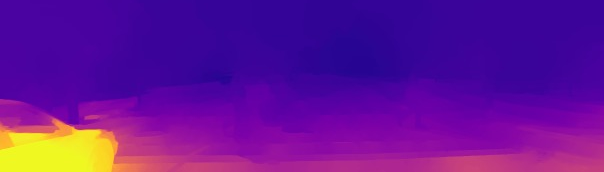}} \\
	\subfloat{\includegraphics[width=.24\linewidth]{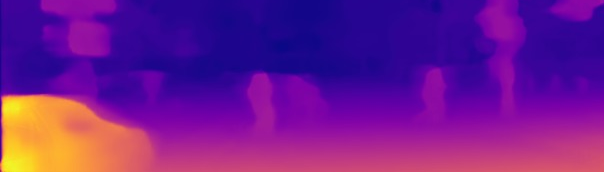}} 
	\hspace{1pt}
	\subfloat{\includegraphics[width=.24\linewidth]{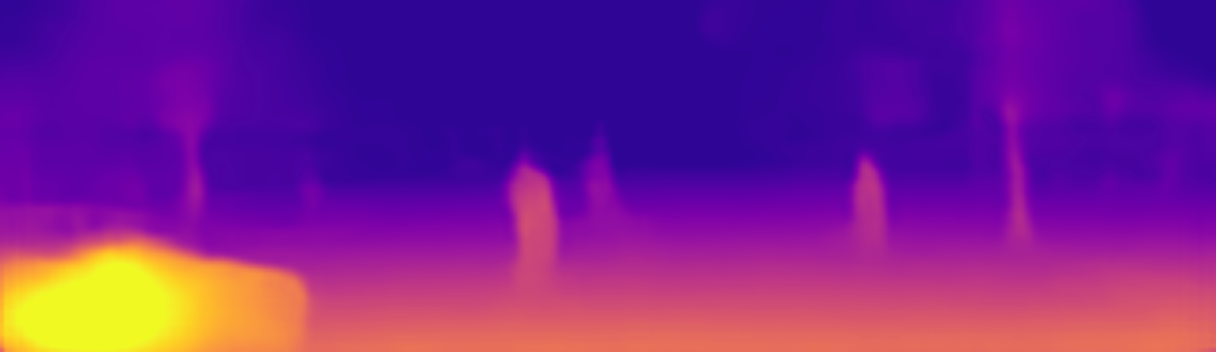}} 
	\hspace{1pt}
	\subfloat{\includegraphics[width=.24\linewidth]{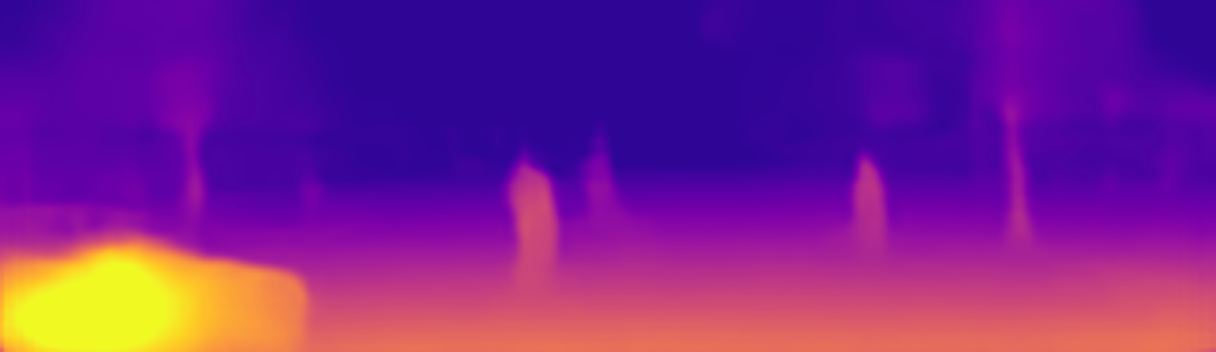}} 
	\hspace{1pt}
	\subfloat{\includegraphics[width=.24\linewidth]{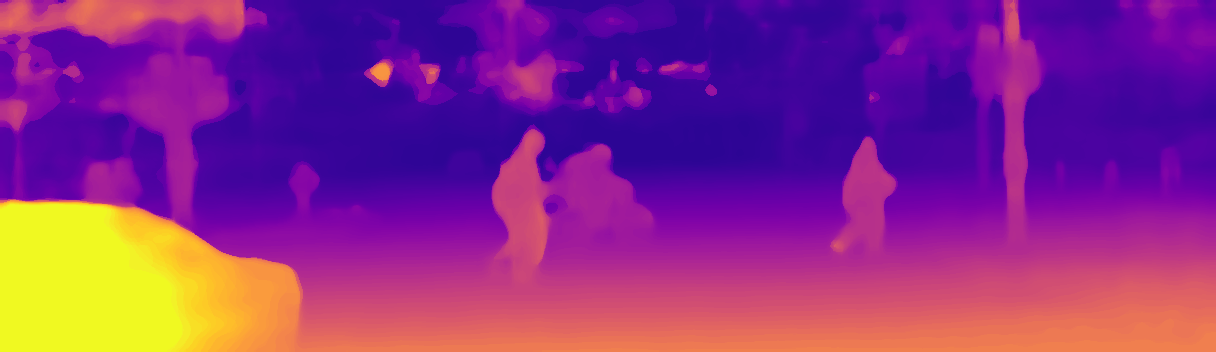}}
	\caption{Qualitative comparison with other state-of-the-art methods on KITTI Eigen test dataset. The ground truth depth maps are interpolated from the sparse LIDAR measurements for better visualization. Our approach can retrieve the depths for small objekts and fine structures better than other methods.}	
	\label{fig:experiments:kitti_comparison}
\end{figure}

\begin{figure}[ht]
	\centering
	\includegraphics[width=.32\linewidth]{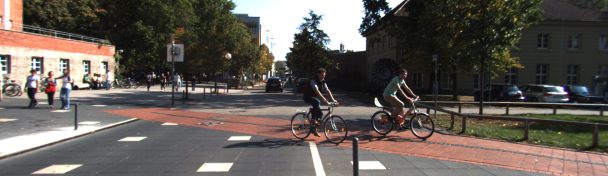} 
	\includegraphics[width=.32\linewidth]{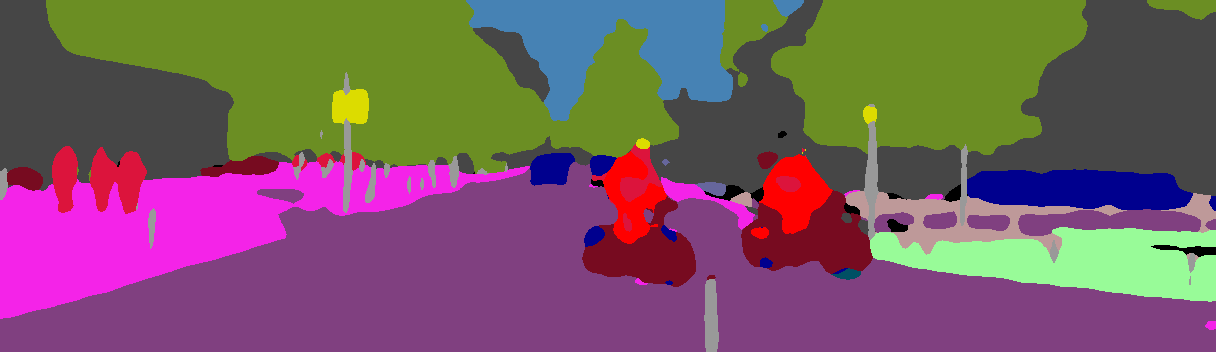} 
	\includegraphics[width=.32\linewidth]{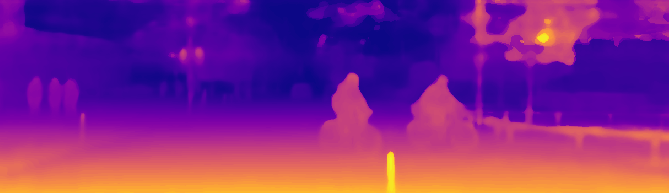} \\
	\vspace{0.1cm}
	\includegraphics[width=.32\linewidth]{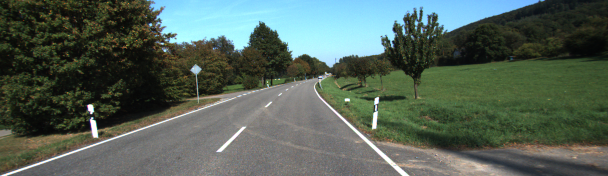} 
	\includegraphics[width=.32\linewidth]{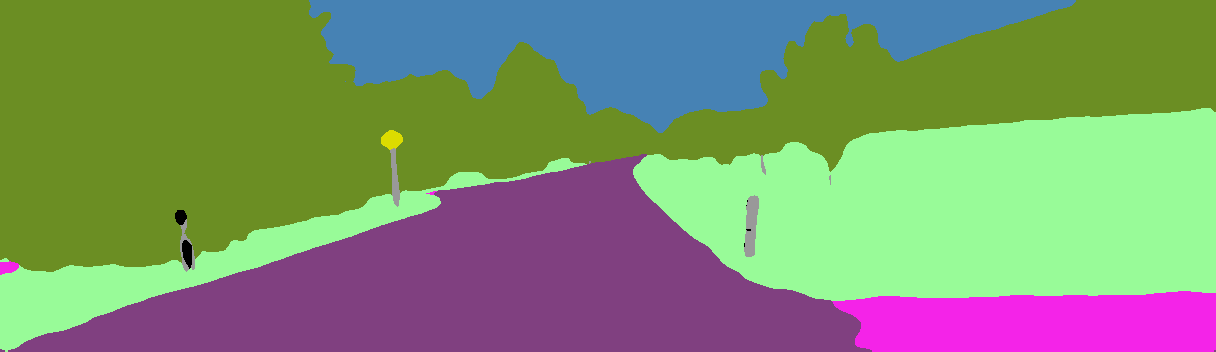} 
	\includegraphics[width=.32\linewidth]{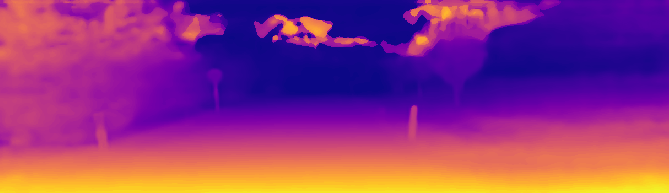} \\
	\vspace{0.1cm}
	\includegraphics[width=.32\linewidth]{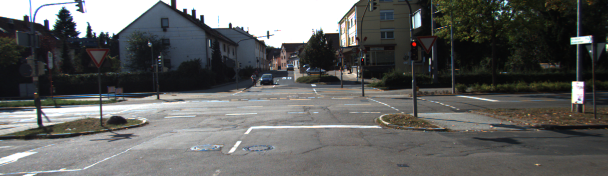} 
	\includegraphics[width=.32\linewidth]{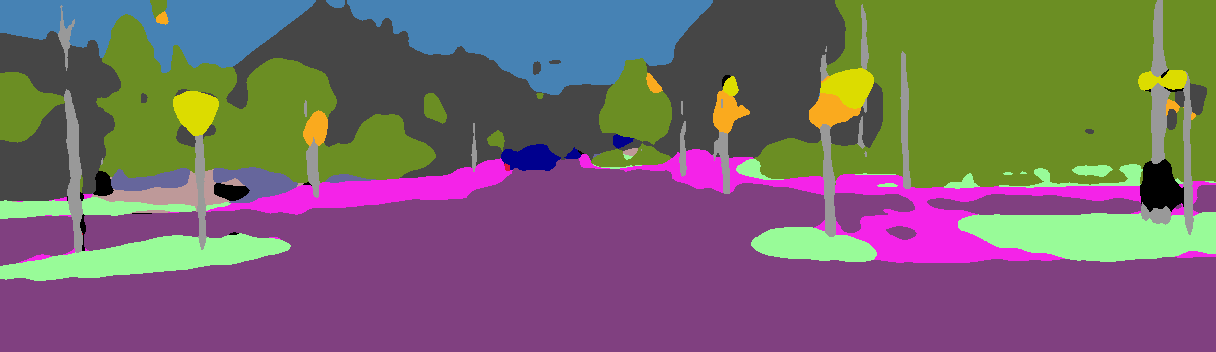} 
	\includegraphics[width=.32\linewidth]{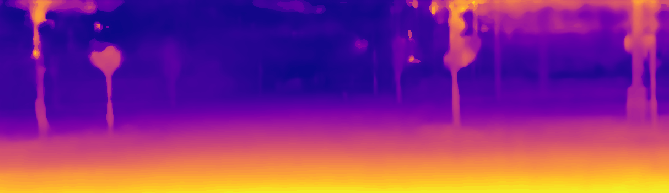} \\
	\vspace{0.1cm}
	\includegraphics[width=.32\linewidth]{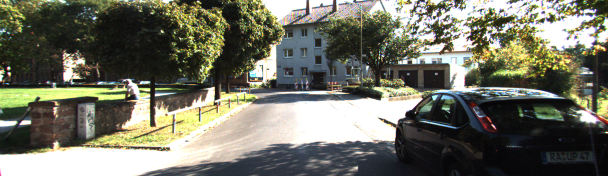} 
	\includegraphics[width=.32\linewidth]{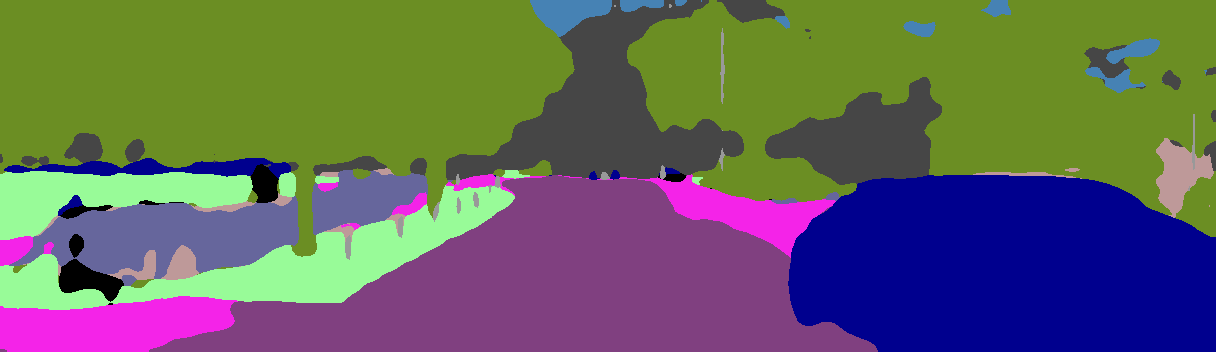} 
	\includegraphics[width=.32\linewidth]{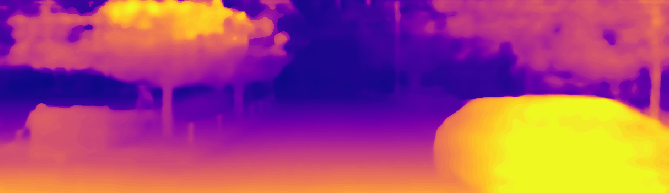} \\
	\caption{Qualitative results of semantic segmentation and depth estimation using examples from KS test dataset.}
	\label{fig:experiments:kitti_semantics}
\end{figure}

The Eigen test dataset consists of $ 697 $ images, which are selected from $ 29 $ different sequences of the KITTI raw dataset. The evaluation in table~\ref{table:experiments:kitti_eigen_split} proves that SDNet achieves very good results in all metrics and is only outperformed by DORN. Moreover, the qualitative comparison of the predicted depth maps in figure~\ref{fig:experiments:kitti_comparison} shows that SDNet can reproduce the depths more accurately and with more details compared to other methods, whose depth maps usually look more blurred. Further examples of depth estimation on KS can be found in figure~\ref{fig:experiments:kitti_semantics}.
\begin{table}[ht]
	\begin{center}
		\begin{tabular}{l | P{1.7cm} P{1.3cm} P{1.3cm} P{1.3cm}}
			Method & SiLog \scriptsize{[$ 100 \cdot \log(\text{m}) $]} & SRD \newline \scriptsize{in \%} & ARD \newline \scriptsize{in \%} & iRMSE \scriptsize{[$1$/km]} \\ 
			\hline
			\hline
			DORN \cite{Fu2018} & 11.67 & 2.21 & 9.04 & 12.23 \\
			VGG16-UNet \cite{Guo2018}  & 13.41 & 2.86 & 10.06 & 15.06 \\
			DABC \cite{Li2018} & 14.49 & 4.08 & 12.72 & 15.53 \\
			\textbf{SDNet} & 14.68 & 3.90 & 12.31 & 15.96 \\
			APMoE \cite{Kong2019} & 14.74 & 3.88 & 11.74 & 15.63 \\
			CSWE \cite{Li2018a} & 14.85 & 3.48 & 11.84 & 16.38 \\
			DHGRL \cite{Zhang2018} & 15.47 & 4.04 & 12.52 & 15.72 \\
			\hline
		\end{tabular}
	\end{center}
	\caption{Official results of the KITTI depth prediction benchmark \cite{Uhrig2017} (as of May 01, 2019). This table includes only already published methods. If several results of one method are ranked, then only the best is taken.}
	\label{table:experiments:kitti_depth_benchmark}
\end{table}

\begin{figure}[ht]
	\centering
	\includegraphics[width=.32\linewidth]{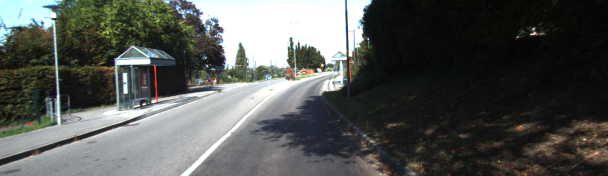} 
	\includegraphics[width=.32\linewidth]{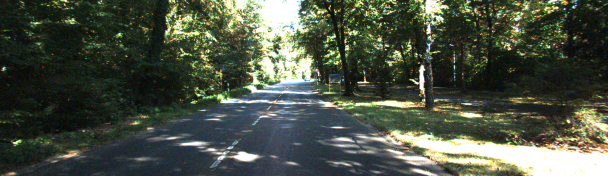} 
	\includegraphics[width=.32\linewidth]{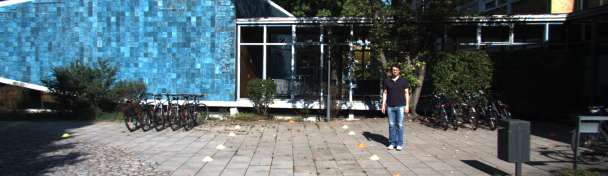} \\
	\vspace{0.1cm}
	\includegraphics[width=.32\linewidth]{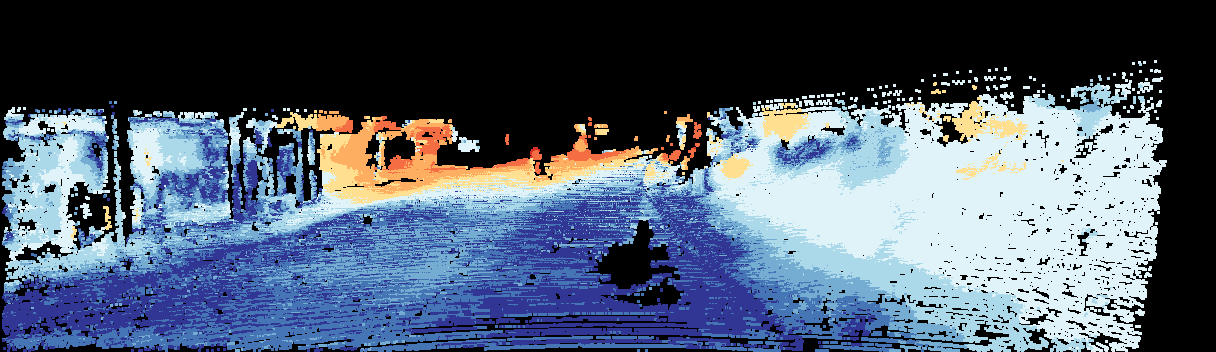}
	\includegraphics[width=.32\linewidth]{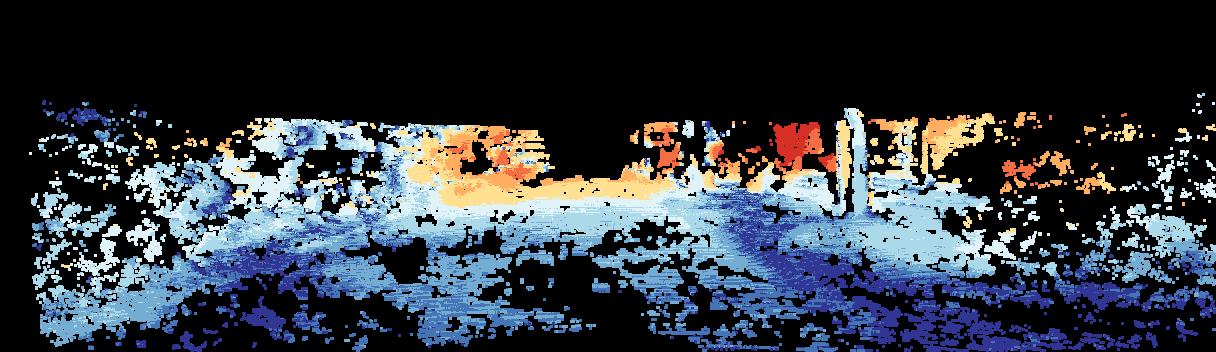}
	\includegraphics[width=.32\linewidth]{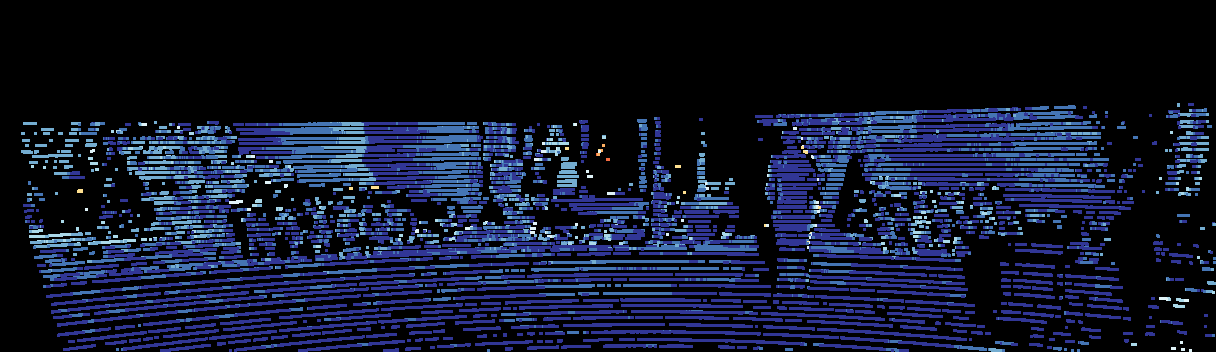}
	\caption{D1 error of the SDNet depth estimate from the ground truth on three exemplary images from the KITTI Depth Prediction test dataset. The errors were computed by KITTI and blue represents a small deviation and red a large one.}
	\label{fig:experiments:kitti_error}
\end{figure}

Besides to this unofficial analysis, the estimated depths of SDNet were also evaluated on the official KD benchmark, which consists of $ 500 $ test images. The images depict different driving situations, so that the network has to generalize in order to perform well in this benchmark. The results are shown in table~\ref{table:experiments:kitti_depth_benchmark}, where only previously published methods have been listed and, again, SDNet achieves state-of-the-art results. Based on the D1 error images in figure~\ref{fig:experiments:kitti_error}, which shows the deviation to the ground truth for some test images, it can be recognized that the predicted depth of SDNet is erroneous especially at large depths. One reason for this could be that the maximum depth class has been set to $ 80 $ m. This also explains the relatively small gap to DORN at the Eigen evaluation, in which the maximum depth is limited to $ 50 $ m or $ 80 $ m.

Additionally, the predicted semantic segmentation of SDNet was also evaluated on the KS benchmark, which consists of $ 200 $ images. The results obtained in this benchmark can be found in table~\ref{table:experiments:kitti_depth_semantic}. Although, the focus of SDNet is the estimation of depth and in consequence the semantic segmentation has not been purposefully optimized and has only been used to learn more meaningful features, SDNet also achieves state-of-the-art results in this benchmark. 

\begin{table}[ht]
	\begin{center}
		\begin{tabular}{l | P{1.5cm} P{1.5cm} P{1.5cm} P{1.5cm}}
			Method & IoU class & iIoU class & IoU cat. & iIoU cat. \\
			\hline
			\hline
			SegStereo \cite{Yang2018a} & 59.10 & 28.00 & 81.31 & 60.26 \\
			\textbf{SDNet} & 51.14 & 17.74 & 79.62 & 50.45 \\
			APMoE \cite{Kong2019} & 47.96 & 17.86 & 78.11 & 49.17 \\
			\hline
		\end{tabular}
	\end{center}
	\caption{Official results of the KITTI pixel-level semantic segmentation benchmark \cite{Alhaija2018} (as of May 01, 2019). This list includes only already published methods.}
	\label{table:experiments:kitti_depth_semantic}
\end{table}

In figure~\ref{fig:experiments:kitti_semantics}, sample results are shown on this test dataset. It can be seen that the results of semantic segmentation and depth estimation correlate directly with each other. For example, if a pixel is assigned to a wrong semantic class, then mostly the depth prediction is also incorrect. If, on the other hand, the depth or semantic class has been correctly classified, then in most cases the other output is correct as well. 

\subsection{Cityscapes Dataset}

\begin{figure}[ht]
	\centering
	\includegraphics[width=.24\linewidth]{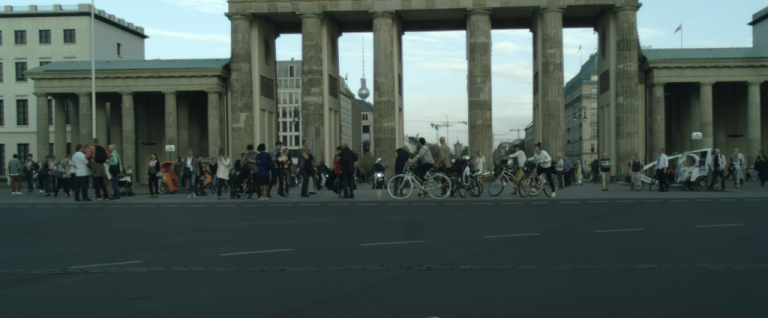} 
	\includegraphics[width=.24\linewidth]{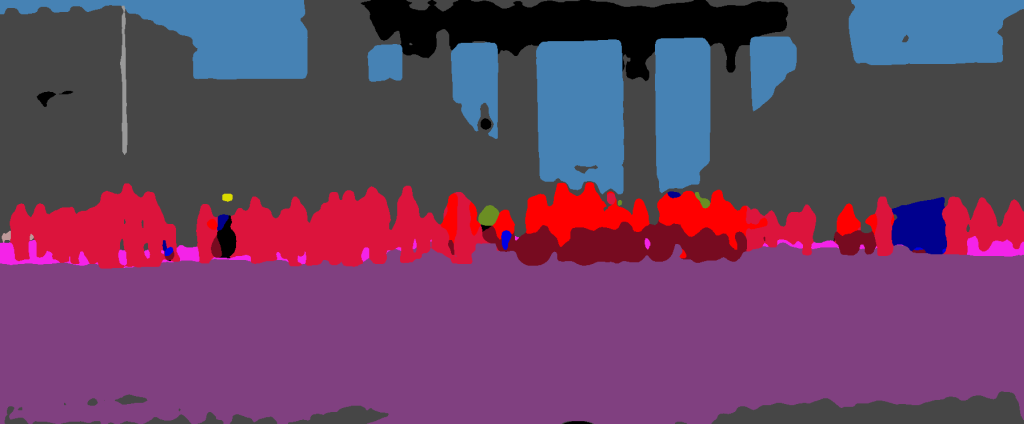}
	\includegraphics[width=.24\linewidth]{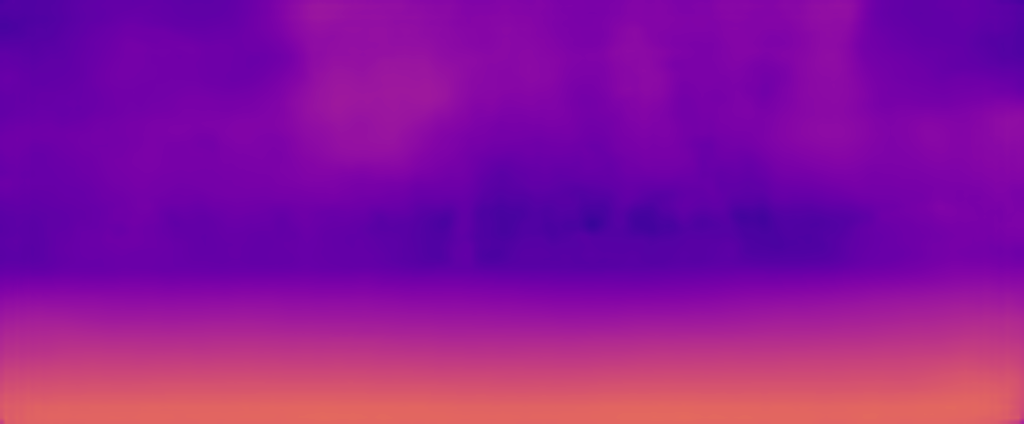}
	\includegraphics[width=.24\linewidth]{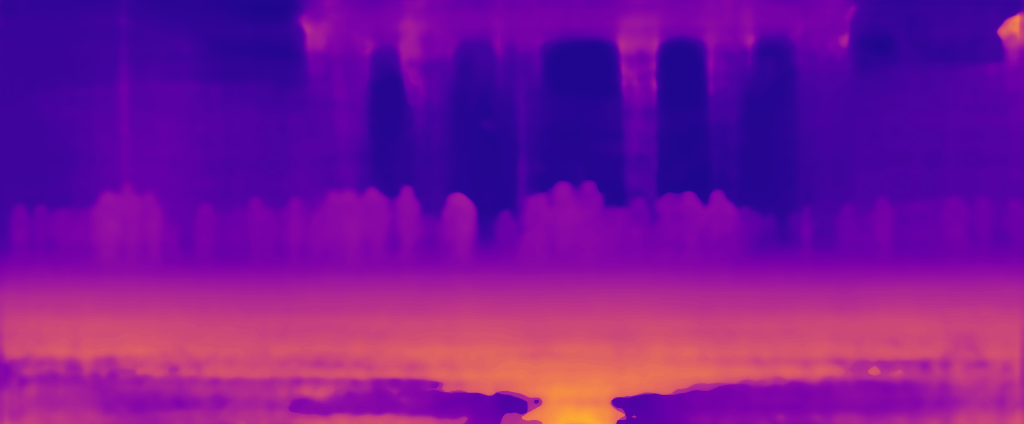} \\
	\vspace{0.1cm}
	\includegraphics[width=.24\linewidth]{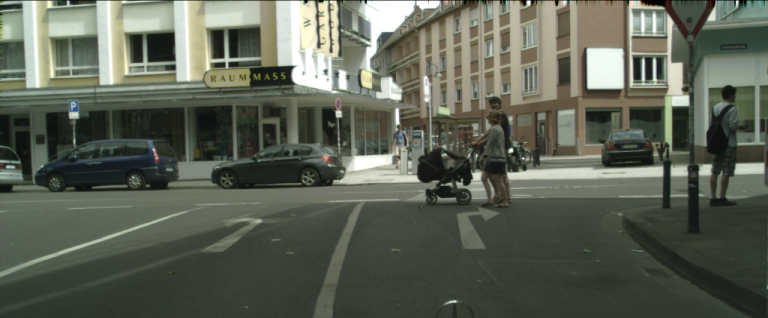} 
	\includegraphics[width=.24\linewidth]{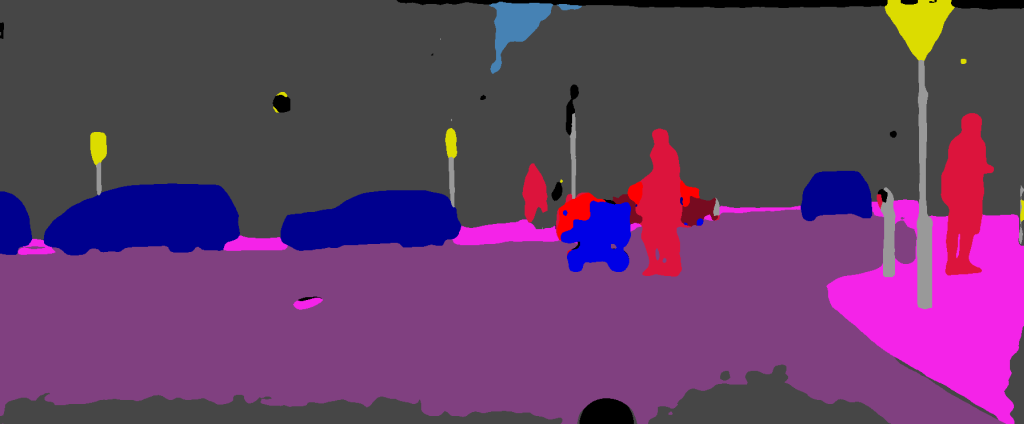}	\includegraphics[width=.24\linewidth]{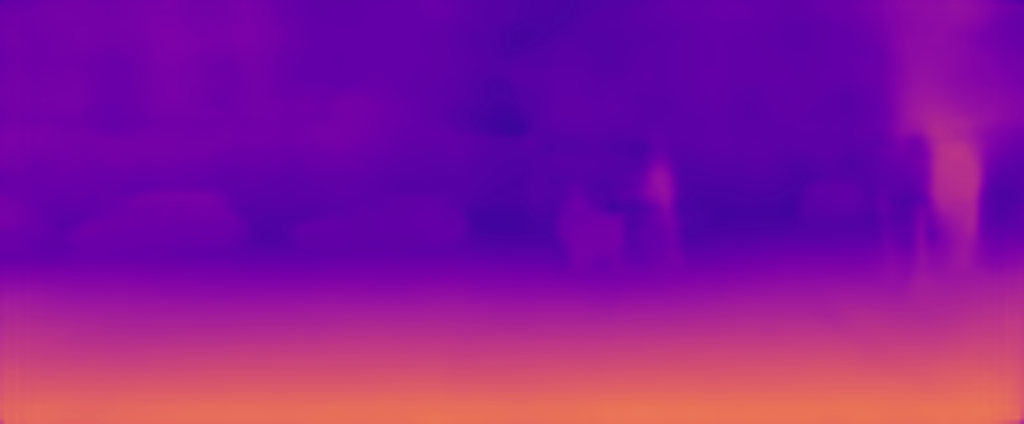}
	\includegraphics[width=.24\linewidth]{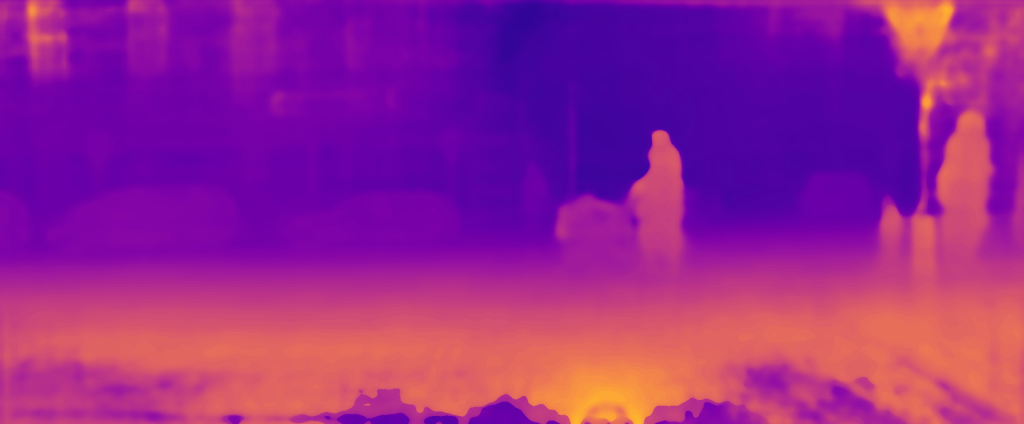} \\
	\vspace{0.1cm}
	\includegraphics[width=.24\linewidth]{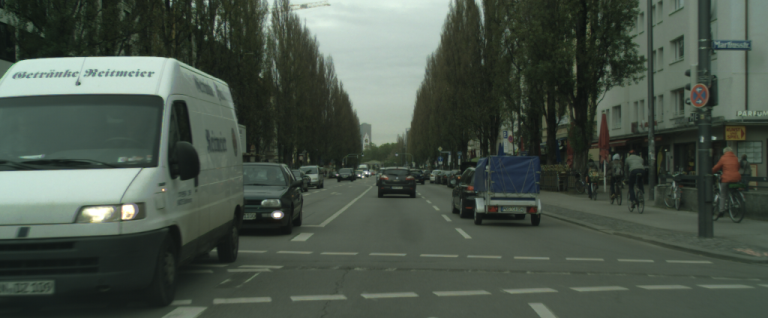} 
	\includegraphics[width=.24\linewidth]{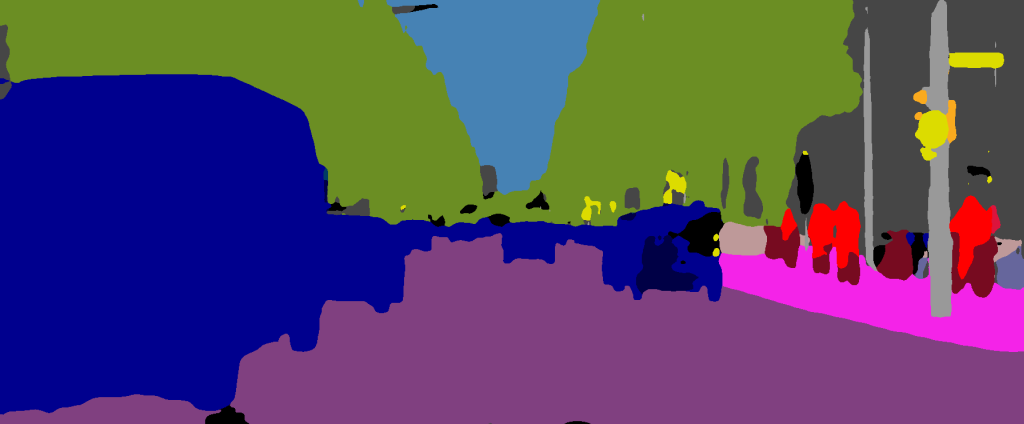} 	\includegraphics[width=.24\linewidth]{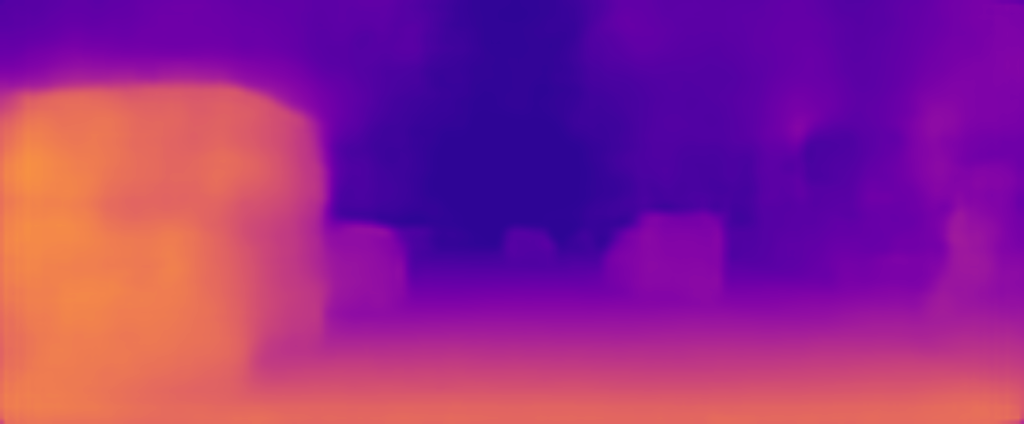}
	\includegraphics[width=.24\linewidth]{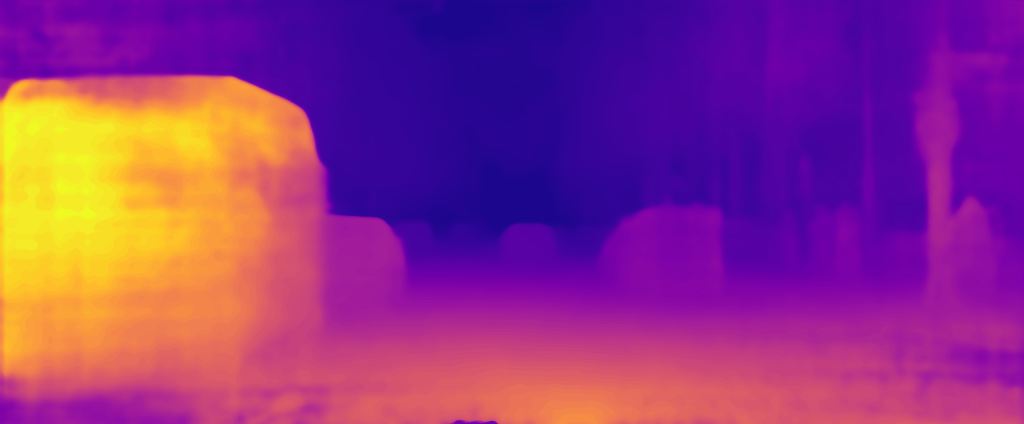} \\	
	\caption{Qualitative results of the semantic segmentation and the depth estimation based on exemplary examples from the CS test dataset. The first column depict the input image and the second the predicted semantics. In the last two columns, the depth maps of \cite{Kuznietsov2017} (third column) and SDNet (last column) are shown.}
	\label{fig:experiments:cityscapes}
\end{figure}

Both outputs of SDNet were also evaluated on CS to show the generalization capability of SDNet. Unfortunately, there is no benchmark for depth estimation, which is why the results could only be assessed qualitatively. In figure~\ref{fig:experiments:cityscapes}, the estimated depth maps of SDNet and \cite{Kuznietsov2017} are exemplarily shown for CS. One can clearly see that the SDNet depths are more detailed compared to Kuznietsov \etal. For example, the depth for the persons in front of the Brandenburg Gate and the gate itself are retrieved well by SDNet in contrast to \cite{Kuznietsov2017}, where the depths are noticeably more blurry. Aside from the depths, the semantic segmentation of SDNet is also shown in this figure. Qualitatively, the predicted semantics yield correct and good results, with a few exceptions. This can also be shown by the quantitative results on CS (see supplementary materials).

\subsection{Ablation Studies}

We also perform ablation studies on the KD validation data to empirically prove that our contributions/enhancements result in verifiable improvements. These experiments can be divided into three parts: i.e. loss function, discretization levels and architecture. The results are shown in table~\ref{table:experiments:ablation_studies}. As a baseline configuration for SDNet, a ResNet-50 with group normalization layers is chosen as encoder and the proposed ordinal BCE loss is used as loss function from equation \ref{eq:approach:bce_loss}. Thus, the CNN is trained only on the depths and the semantic loss is completely ignored. The depth interval $ [2 \text{m}; 80 \text{m}] $ is divided into $ M_d = 128 $ classes. 

\begin{table}[ht]
	\begin{center}
		\begin{tabular}{l | P{1.7cm} P{1.3cm} P{1.3cm} P{1.3cm} P{1.3cm}}
			Method & SiLog \scriptsize{[$ 100 \cdot \log(\text{m}) $]} & ARD \newline \scriptsize{in \%} & RMSE \newline \scriptsize{[m]} & iRMSE \scriptsize{[$1$/km]} & Accuracy \newline \scriptsize{in \%}  \\ 
			\hline
			\hline
			sem. loss $(\lambda = 10)$ & 14.83 & 9.39 & 3.11 & 12.17 & 20.10 \\
			MAE & 18.80 & 12.11 & 4.03 & 15.30 & - \\ 
			\hline
			$ M_{\text{depth}} =  96$ & 16.04 & 10.88 & 3.30 & 13.28 & 22.50 \\
			$ M_{\text{depth}} =  160$ & 15.57 & 10.06 & 3.23 & 12.87 & 15.50 \\
			\hline
			BatchNorm & 15.84 & 10.31 & 3.42 & 12.77 & 18.95 \\ 
			ResNet-101 & 16.25 & 10.89 & 3.44 & 13.12 & 18.24 \\ 
			\hline
			baseline & 15.38 & 9.92 & 3.30 & 12.55 & 20.50 \\ 
			\hline
		\end{tabular}
	\end{center}
	\caption{Ablation studies on the KD validation dataset for different configurations. As baseline, we use SDNet without semantic loss $(\lambda = 0)$.}
	\label{table:experiments:ablation_studies}
\end{table}

The performance of the network can be increased, if the semantic loss is added to the loss. Hence, the network can learn superior features that do not solely depend on the depth, but also on the semantics. As a result, as shown here empirically, a better performance can be achieved. In addition, the ordered classification of the depths is superior to a regression approach with MAE.

The depth estimation achieves the best results, when dividing the depth interval into $ 128 $ classes. Although the accuracy is larger for less classes, the prediction results get worse, because the discretization error increases with fewer number of classes. On this basis, $ 128 $  classes have proven to be the best trade-off.

Replacing the ResNet-50 by a ResNet-101 as an encoder, was supposed to increase performance, because even better features can be extracted. However, the evaluation shows that exactly the opposite was the case. The reasons for this could be that larger networks generally are harder to train. Therefore, adjustments or fine-tuning had to be made in the training protocol, which was not done (e.g. additional warm-up phase or adjustment of the learning rate). Additionally, the change of the normalization method was also evaluated. By default, batch normalization layers are used in the ResNets, but these proved disadvantageous for small batch sizes. Therefore, these layers have been replaced by group normalization layers, which had a positive effect on all metrics.
\section{Summary \& Conclusion}

In this paper, we propose the novel SDNet for the simultaneous prediction of pixel-wise depths and semantic labels from a single monocular image. The architecture of SDNet is based on the DeepLabv3+ model, which we have extended. The depths are determined by ordered classes rather than the classic regression-based approach. In comparison to other methods, the input image is also segmented semantically. These two modifications turn out to be beneficial as the CNN learns semantically richer features, resulting in significant better results. 

In a further stage, a new error or uncertainty measure could be developed on basis of the output, because prediction errors occur mostly in both predictions. This knowledge could be used for such a measure. Another improvement of SDNet would be that a better encoder than ResNet-50 is used. But this implies also that hyper-parameters must be tuned during training, which by itself could be an improvement.  Nevertheless, SDNet yields state-of-the-art results for both semantic segmentation and depth estimation on various datasets.

\subsubsection*{Acknowledgement:}

This project (HA project no. 626/18-49) is financed with funds of LOEWE – Landes-Offensive zur Entwicklung Wissenschaftlich-öko- nomischer Exzellenz, Förderlinie 3: KMU-Verbundvorhaben (State Offensive for the Development of Scientific and Economic Excellence).

We also gratefully acknowledge the support of NVIDIA Corporation with the donation of the Titan Xp GPU used for this research.

\clearpage
\bibliographystyle{splncs04}
\bibliography{gcpr}

\end{document}